\documentclass[12pt, titlepage, reqno]{article}

 \usepackage[cmex10]{amsmath}

 \usepackage{amssymb, latexsym}
 \usepackage{amsmath} 
 \usepackage{amsthm}
 \usepackage{bm}
 \usepackage[mathscr]{eucal}
 \usepackage{enumerate}

 \usepackage{natbib}
 \usepackage[french,english]{babel}

\usepackage{algorithm,algpseudocode}
\usepackage{caption}
\usepackage{color}

\usepackage{gensymb}

\usepackage[caption = false]{subfig}

\usepackage{float}

\usepackage{algpseudocode}

\usepackage{graphicx}

\usepackage{tabularx}
\usepackage{slashbox}

\usepackage{hyperref}
\hypersetup{
    colorlinks,%
    citecolor=blue,%
    filecolor=blue,%
    linkcolor=blue,%
    urlcolor=blue,
}  
  
\usepackage{multirow}

\graphicspath{{./Figures/}}

\begin{document}

 \begin{titlepage}

 \begin{center}

\textbf{Particle Filtering for PLCA model with Application to Music Transcription}

\vspace{10ex}

Cazau Dorian$^{a}$ \footnote{Corresponding author e-mail: cazaudorian@outlook.fr}, Guillaume Revillon$^{a}$, Yuancheng Wang$^{a}$ and Olivier Adam$^{a}$\\ 

\vspace{1ex}

\begin{flushleft}

\small{$^a$ Sorbonne Universit\'es, UPMC University Paris 06/CNRS, UMR 7190, Institut Jean le Rond, d'Alembert, F-75015, Paris, France} \\

\end{flushleft}

 \end{center}

 \end{titlepage}

\begin{abstract}
Automatic Music Transcription (AMT) consists in automatically estimating the notes in an audio recording, through three attributes: onset time, duration and pitch. Probabilistic Latent Component Analysis (PLCA) has become very popular for this task. PLCA is a spectrogram factorization method, able to model a magnitude spectrogram as a linear combination of spectral vectors from a dictionary. Such methods use the Expectation-Maximization (EM) algorithm to estimate the parameters of the acoustic model. This algorithm presents well-known inherent defaults (local convergence, initialization dependency), making EM-based systems limited in their applications to AMT, particularly in regards to the mathematical form and number of priors. To overcome such limits, we propose in this paper to employ a different estimation framework based on Particle Filtering (PF), which consists in sampling the posterior distribution over larger parameter ranges. This framework proves to be more robust in parameter estimation, more flexible and unifying in the integration of prior knowledge in the system. Note-level transcription accuracies of 61.8 $\%$ and 59.5 $\%$ were achieved on evaluation sound datasets of two different instrument repertoires, including the classical piano (from MAPS dataset) and the marovany zither, and direct comparisons to previous PLCA-based approaches are provided. Steps for further development are also outlined.
\end{abstract}

\addtocounter{page}{2}

\section{Introduction}

\subsection{Background on PLCA}

Probabilistic Latent Component Analysis (PLCA) is a straightforward extension of Probabilistic Latent Semantic Indexing \citep{Hofmann1999} which deals with an arbitrary number of dimensions and can exhibit various features such as sparsity or shift-invariance. The basic model is defined as

\begin{equation}\label{GenePLCA}
P(x) = \sum_z P(z) \prod_{j=1}^J P(x_j | z)
\end{equation}

where $P(x)$ is an $J$-dimensional distribution of the random variable $x=(x_1,\ldots,x_J)$, $z$ is a latent variable and the $P(x_j|z)$ are one dimensional distributions with $j\in\{1,\ldots,J\}$. Such a general model has been successfully applied to audio signals, with a theoretical framework developed by \citep{Smaragdis2006}. Especially, PLCA has been proven to be an efficient probabilistic tool for non-negative data analysis, which offers a convenient way of designing spectrogram models. From its general formulation (eq. \ref{GenePLCA}), and considering a spectrogram $S(f,t)$ as a probability distribution $P(f,t)$, a latent variable $z$ is introduced to model $P(f,t)$ as

\begin{equation}\label{PLCAaudio}
	P(f,t)=\sum_z P(z)P(f|z)P(t|z)=\sum_z P(z,t)P(f|z)
\end{equation}

where $f$ and $t$ represent respectively frequency and time, and are both conditionally independent given $z$, $P(f|z)$ are the spectral bases corresponding to component $z$, and $P(z,t)$ their time activations. Since there is usually no closed-form solution for the maximization of the log-likelihood or the posterior, iterative update rules based on the Expectation-Maximization (EM) algorithm are employed to estimate $P(f|z)$ and $P(z,t)$.

\subsection{Limitations of current PLCA models}

The major limitation of current PLCA models lies in the inherent problems of the EM algorithm. This algorithm was originally introduced by \citep{Dempster1977} to overcome the difficulties in maximizing likelihoods of missing data models. The main advantage of that method is its easy implementation, consisting of initializing the parameters and iterating expectation and maximization likelihoods in a step-by-step process until convergence. Its major drawback, besides the requirement of convex likelihoods, lies in its sensitivity to initialization, which increase the risks to local convergences \citep{Robert1999}. That issue is exacerbated in the case of multimodal likelihoods. Indeed, the increase of the likelihood function at each step of the algorithm ensures its convergence to the maximum likelihood estimator in the case of unimodal likelihoods, but implies a dependence on initial conditions for multimodal likelihoods. Alternative techniques have also been proposed to optimize the search of global maxima, such as running the algorithm a number of times with different, random starting points, or using variants from the basic EM algorithm such as Deterministic Annealing EM (DAEM) algorithm \citep{Ueda1998}. These theoretical issues have reached research fields working on audio signals. To tackle the problem of dependency to initialization, some authors \citep{Grindlay2010,Benetos2013} perform a training of the instrument templates, which has proved to be an effective way to initialise the spectral bases. Indeed, by fixing them without data-driven updating, we obtain a stable output for the gain function, independent of its initialisation. However, when the model becomes more complex with for example the introduction of different instrument variables, performing robust initialization is more difficult. For what concerns the local convergence problem, some works \citep{Hoffman2009,Grindlay2010,Cheng2013} have used the DAEM algorithm based on a temperature parameter.

This limitation becomes particularly critical when integrating priors into the PLCA framework. Generally speaking, this integration introduces generic problems in optimization convergence to global maxima, especially when the prior has a multi-modality form. Indeed, when a prior is injected, the maximization step becomes a maximum a posteriori step and the log posterior probability needs to have the right properties for maximization. \citep{Fuentes2013} used of a numerical fixed point algorithm to solve the modified EM equations with a sparsity prior, whose convergence is only theoretically supposed, but "observed in practice" (although the sensitivity of the algorithm convergence to the evaluation sound dataset is not detailed). \citep{Benetos2013} privileged the use of pre-defined templates, which allows them to skip computing the EM update equation of templates, and just to apply a sparsity constraint on the pitch activity matrix and the pitch-wise source contribution matrix. Also, the simultaneous use of several priors on a same model parameter leads to some difficulties in terms of mathematical calculation and increases convergence problems \citep{Fuentes2013}.

\subsection{Particle filtering}

In the framework of Bayesian variable selection, Markov Chain Monte Carlo, or Particle filtering (PF), type approaches have been proposed \citep{Fevotte2006b,Fevotte2008}. These methods consist in sampling the posterior distribution over larger parameter ranges, making them more demanding than their EM-like counterparts, but which also, in return, offer increased robustness in convergence (i.e. reduced problems of convergence to local minima) and a complete Monte Carlo description of this parameter posterior density \citep{Fevotte2006,Fevotte2006b,Fevotte2008}.

\subsubsection{General Overview}

Many problems in statistical signal processing \citep{Fong2002,Andrieu2003,Vermaak2000} can be stated in  a state space form as follows,

\begin{equation}\label{statespace}
	x_{t+1}\sim f(x_{t+1}|x_t) 
\end{equation}	
\begin{equation}	
	y_{t+1}\sim g(y_{t+1}|x_{t+1}) 
\end{equation}

where $\{x_t\}$ are unobserved states of the system and $\{y_t\}$ are observations made over some time, $t$. $f(.|.)$ and $g(.|.)$ are pre-specified state evolution and observation densities. A primary concern in many state-space inference problems is the sequential estimation of the filtering distribution $p(x_t|y_{1:t})$, and the simulation of the entire smoothing distribution $p(x_{1:t}|y_{1:t})$, where $y_{1:t}=({y_1,y_2,\cdots,y_t})$ and $x_{1:t}=({x_1,x_2,\cdots,x_t})$. Updating of the filtering distribution can be achieved, in principle, using the standard filtering recursions \citep{Robert1999}  

\begin{equation}\label{filt}
	p(x_{t+1}|y_{1:t})=\int{p(x_t|y_{1:t})f(x_{t+1}|x_t)dx_t}
\end{equation}
\begin{equation}
	p(x_{t+1}|y_{1:t+1})=\frac{g(y_{t+1}|x_{t+1})p(x_{t+1}|y_{1:t})}{p(y_{t+1}|y_{1:t})}
\end{equation}

Smoothing can also be performed recursively backwards in time using the smoothing formula \citep{Robert1999}  

\begin{equation}\label{smooth}
	p(x_{t}|y_{1:T})=\int{p(x_{t+1}|y_{1:T})\frac{p(x_t|y_{1:t})f(x_{t+1}|x_t)}{p(x_{t+1}|y_{1:t})}dx_t}
\end{equation}

In practice, these filtering (eq. \ref{filt}) and smoothing (eq. \ref{smooth}) computations can only be performed in closed form for linear Gaussian models using the Kalman filter / smoother, and for finite state-space hidden Markov models. In the case of non-linear non-Gaussian models, there is no general analytic expression for the computations of these density functions. As a consequence, an approximation strategy is required to estimate the filtering and smoothing densities, which is commonly performed with the PF method, also known as sequential Monte Carlo methods. Within the PF framework, the filtering distribution is approximated with an empirical distribution formed from point masses also called particles,

\begin{equation}\label{ApproxPF}
	p(x_t|y_{1:t})\approx \sum\limits_{i=1}^Nw_t^{(i)}\delta(x_t-x_t^{(i)})
\end{equation}	
\begin{equation}
	\sum\limits_{i=1}^Nw_t^{(i)}=1,w_t^{(i)}\ge 0
\end{equation}

where $\delta(.)$ is the Dirac delta function and $w_t^{(i)}$ is a weight attached to particle $x_t^{(i)}$. Given this particle approximation to the posterior distribution, we can estimate the expected value of any function $f$ $w.r.t$ the distribution $I(f,t)$, defined as $I(f_t)=\int f(x_t)p(x_t|y_{1:t})dx_t$, using the following Monte Carlo approximation

\begin{equation}
	I(f_t)\approx \sum\limits_{i=1}^Nf(x_t^{(i)})w_t{(i)}
\end{equation}

Particle smoothers generate batched realisations of $p(x_{1:T}|y_{1:T})$ based on the forward PF results. In other words, the particle smoothers are an efficient method for generating realisations from the entire smoothing density $p(x_{1:T}|y_{1:T})$ using filtering approximation.

\subsubsection{Filtering}
We consider the filtering distribution $p(x_t|y_{1:t})$. Using the Bayes' rule, this distribution can be rewritten as follows,

\vspace{-0.5cm}
\begin{align}
	p(x_t|y_{1:t}) &=p(x_t|y_t,y_{1:t-1})\\
			&\propto p(y_t|x_t,y_{1:t-1})p(x_t|y_{1:t-1})\\
			&\propto g(y_t|x_t)p(x_t|y_{1:t-1})\\
			&\propto \int g(y_t|x_t)f(x_t|x_{t-1})p(x_{1:t-1}|y_{1:t-1})dx_{1:t-1}	
\end{align}

Assuming that a particle approximation to $p(x_{1:t-1}|y_{1:t-1})$ has already been generated,

\begin{equation}
	p(x_{1:t-1}|y_{1:t-1})\approx\sum\limits_{i=1}^N\delta(x_{1:t-1}-x_{1:t-1}^{(i)}) 
\end{equation}

Then, assuming that $f(x_t|x_{t-1})$ and $g(y_t|x_t)$ can be evaluated pointwise, we generate, for each state trajectory $x_{1:t-1}^{(i)}$, a random sample from a proposal distribution $q(x_t|x_{1:t-1}^{(i)},y_{1:t})$. Then, the weights $w_t$ of the filtering distribution (eq. \ref{ApproxPF}) can be approximated by 

\begin{equation}
	w_t^{(i)}\approx\frac{g(y_t|x_t^{(i)})f(x_t^{(i)}|x_{t-1}^{(i)})}{q(x_t^{(i)}|x_{1:t-1}^{(i)},y_{1:t})}
\end{equation}

Finally, we perform a multinomial resampling step, such that the probability that $x_t^{(i)}$ is selected is proportional to $w_t^{(i)}$, to obtain an unweighted approximate random draw from the filtering distribution $p(x_t|y_{1:t})$. It is noteworthy that if the resampling step is forgotten, a degeneracy phenomenon can occur. Indeed, after a few iterations, all but one particle will have negligible weight. \citep{Doucet2000} has shown that the variance of the importance weights can only increase over time, and thus, it is impossible to avoid the degeneracy phenomenon. This degeneracy implies that a large computational effort is devoted to updating particles whose contribution is almost zero. As a result, a resampling step is needed to eliminate particles with small weights and generate a new set  $\{x_t^{(i)}\}_i$, which is an i.i.d. (independent and identically distributed) sample from the approximate density $p(x_t|y_{1:t})$, with a resetting of the weights $\{w_t^{(i)}\}_i$ to $1/N$.

\subsubsection{Smoothing}

The entire smoothing density $p(x_{1:T}|y_{1:T})$ can be factorized as :

\begin{equation}
	p(x_{1:T}|y_{1:T})=p(x_T|y_{1:T})\prod_{t=1}^{T-1}p(x_t|x_{t+1:T},y_{1:T})
\end{equation}

Using the filter approximation (eq. \ref{ApproxPF}) to $p(x_t|y_{1:t})$ and the Markovian assumptions of the model, we can write,

\vspace{-0.5cm}
\begin{multline}
	p(x_t|x_{t+1:T},y_{1:T}) \propto p(x_t|y_{1:t})f(x_{t+1}|x_t)\\
	\approx \sum\limits_{i=1}^Nw_{t|t+1}^{(i)}\delta(x_t-x_t^{(i)})
\end{multline}

with the modified weights 

\begin{equation}
	w_{t|t+1}^{(i)}=\frac{w_t^{(i)}f(x_{t+1}|x_t^{(i)})}{\sum_{j=1}^Nw_t^{(j)}f(x_{t+1}|x_t^{(j)})}
\end{equation}

This revised particle distribution can be used to generate states successively in the reverse-time direction, conditioning upon future states.

\subsection{Our contributing work}

The main objective of this paper is to propose an alternative formulation of current PLCA models applied to audio signals, replacing the EM algorithm by a more generic parameter estimation algorithm based on a PF method. We call this new algorithm PLCA-PF in the following. The main advantage expected from this new algorithm is to be able to scan the whole parameter space so as to take into account any features of the parameters, and thus overcoming the limitations underlined in our introduction specific to current PLCA models. In regards to prior integration particularly, this new framework allows releasing the constraints on prior mathematical forms and number. This paves the way towards more complete modelings of the multi-faceted information carried by musical signals, covering both time (e.g. tempo and rhythm) and frequency (e.g. note spectra and chords) domains, and the different prior knowledge classes related to musicology, timbre and playing style.

\section{Particle Filtering for PLCA}

\subsection{State space representation}

Considering the equations \ref{GenePLCA} - \ref{PLCAaudio}, the PLCA can be expressed as :

\vspace{-0.5cm}
\begin{multline}
	P(x,t) =\sum_zP(z,t)P(x|z)\\
	=\sum_{z_1,\ldots,z_K}P(z_1,\ldots,z_K,t)\prod \limits_{j=1}^JP(x_j|z_1,\ldots,z_K)\\
	%&=\sum_{z_1,\ldots,z_K}P(z_1|z_2,\ldots,z_K,t)P(z_2,\ldots,z_K,t)\prod \limits_{j=1}^JP(x_j|z_1,\ldots,z_K)\\
	=\sum_{z_1,\ldots,z_K}P(z_K,t)\prod\limits_{k=1}^{K-1}P(z_k|z_{k+1},\ldots,z_K,t) \\ 
	 \prod \limits_{j=1}^JP(x_j|z_1,\ldots,z_K)
	%&=\sum_{z_1,\ldots,z_K}P_t(z_1,\ldots,z_K)\prod \limits_{j=1}^JP(x_j|z_1,\ldots,z_K)
\end{multline}

with : 

\begin{itemize}
	\item $z \in Z_1 \times \ldots \times Z_K$ is a vector of $K$ latent components $(z_1,\ldots,z_K)$ associated to a finite subset $Z_k=\{1,\ldots,L_k\}$
	\item $t\in\{0,\ldots,T\}$ is the time variable  
	\item $x \in X_1 \times \ldots \times X_J$ is a vector of $J$ features $(x_1,\ldots,x_J)$  where $X_j=\{1,\dots,F_j\}$
\end{itemize}

In this decomposition, $P(z_K,t)$ can be seen as the activation distribution of the latent variable $z_K$, $P(z_k|z_{k+1},\ldots,z_K,t)$ as the weight of the variable $z_k$ conditionally to $(z_{k+1},\ldots,z_K)$ and $P(x_j|z_1,\ldots,z_K)$  as the $J$ features basis. \\

To estimate the set of parameters  $p_t=\{P(z_K,t),P(z_k|z_{k+1},\ldots,z_K,t) \forall k\in\{1,\ldots,K-1\}\}$ at each time $t\in\{0,\ldots,T\}$, the model can be rearranged as a state space process 
\begin{equation}
	p_t \sim f(p_t|p_{t-1})
\end{equation}
\begin{equation}	
	y_t \sim g(y_t|p_t)
\end{equation}

where $f$ is the transition state density function for $p_t$ defined above and $g$ the observation function of $y_t$.

\subsection{Transition and observation densities}

%\textcolor{red}{Justify specific use of the transition density, in relation to PLCA or AMT}

\subsubsection{Transition density}\label{transidensity}

%For all $k\in \{1,\ldots,K\}$, $i_k \in card(Z_k)$ denotes the $i_k^{th}$ component of the latent variable $z_k$.

Assuming that each latent variable $z_k\in Z_k$ is i.i.d, each marginal vector $P(z_{K},t)$ and $P(z_k|z_{k+1},\ldots,z_K,t)$ can be independently estimated. Recalling that at a given time $t$, $P(z_{K},t)$ and $P(z_k|z_{k+1},\ldots,z_K,t)$ represent distributions, Dirichlet priors are injected to ensure that their elements belong to $[0,1]$, as follows, $\forall (z_2,\ldots,z_K)\in Z_2\times\ldots\times Z_K,\forall k\in\{1,\ldots,K-1\}$

\begin{equation}
	P(z_{K},t) \sim Dir(\theta_t^1,\ldots,\theta_t^{L_K})
\end{equation}	

\vspace{-0.5cm}
\begin{multline}
P(z_k|z_{k+1},\ldots,z_K,t) \sim Dir(\delta_k(z_{k},\ldots,z_{K-1})_t^1, \\ 
\ldots,\delta_k(z_{k},\ldots,z_{K-1})_t^{L_K})
\end{multline}

where $\theta$ and $\delta_k$ are random variables representing the weight of each component of  $z_K$ in $P(z_{K},t)$ and $P(z_k|z_{k+1},\ldots,z_K,t)$. %conditionally to $(z_2\ldots,z_{K})$. 
That injection leads to the following hierarchical model 

\begin{equation}\label{HierModel}
 	\begin{matrix}
	H_t&\rightarrow&H_{t+1} \\
	\downarrow & &\downarrow \\
	P_t& &P_{t+1}\\
	\downarrow& &\downarrow\\
	Y_t& &Y_{t+1}
	\end{matrix}
\end{equation}

with $H_t=(\Theta_t,\Delta_t)$ the new states defined by

\begin{equation}
\Theta_t =\{\theta_{t}^{z_K},\forall z_K \in Z_K\}
\end{equation}
\begin{equation}
 \Delta_t =\{\delta_k(z_k,\ldots,z_{K-1})_t^{z_K},\forall z_K \in Z_K\}
\end{equation}

where we have defined

\begin{equation}
	\theta_{t+1}^{z_K}=\theta_{t}^{z_K}\times \alpha_t^{z_K},\alpha_t^{z_K}\sim \phi 
\end{equation}

\vspace{-0.5cm}
\begin{multline}
\delta_k(z_k,\ldots,z_{K-1})_{t+1}^{z_K}= \delta_k(z_k,\ldots,z_{K-1})_{t}^{z_K}\times \\ \gamma_t^{z_K},\gamma_t^{z_K} \sim\psi_k 
\end{multline}

with $\phi$ and $\psi_k$ are positive distributions.

	% \Gamma(a_{(z_2,\ldots,z_{K})}^{z_1},b_{(z_2,\ldots,z_{K})}^{z_1}) \\

%\Gamma(c_{(z_2,\ldots,z_{K})}^{z_1},d_{(z_2,\ldots,z_{K})}^{z_1})

%with  hyperparameters $a_{(z_2,\ldots,z_{K})}^{z_1}$ ,$b_{(z_2,\ldots,z_{K})}^{z_1}$, $c_{(z_2,\ldots,z_{K})}^{z_1}$ and $d_{(z_2,\ldots,z_{K})}^{z_1}$.\\

%It can be noted that conditionally to $\theta_t$ and $\delta_t$, 

%\begin{align}\label{gammaprior1}
%	\theta(z_2,\ldots,z_{K})_{t+1}^{z_1}&\sim\Gamma\left(a_{(z_2,\ldots,z_{K})}^{z_1},\frac{b_{(z_2,\ldots,z_{K})}^{z_1}}{\theta(z_2,\ldots,z_{K})_{t}^{z_1}}\right)\\
%	\delta(z_2,\ldots,z_{K})_{t+1}^{z_1}&\sim\Gamma\left(c_{(z_2,\ldots,z_{K})}^{z_1},\frac{d_{(z_2,\ldots,z_{K})}^{z_1}}{\delta(z_2,\ldots,z_{K})_{t}^{z_1}}\right)
%\end{align}

\subsubsection{Observation density}

$y_t$ has been defined as a representation of $x$ at time $t$. In that state space approach, each component of $y_t$ is represented by the sum of the PLCA model and a white noise, $\forall x \in X_1 \times \ldots \times X_J$,

\vspace{-0.5cm}
\begin{multline}
	y_t(x) =P(x,t)+V_t =\sum_{z_1,\ldots,z_K}P_t(z_1,\ldots,z_K) \\
	\prod \limits_{j=1}^JP_t(x_j|z_1,\ldots,z_K)+V_t
\end{multline}

where $V_t\sim N(0,\sigma^2)$. Denoting $\hat{y_t}$ the vector of components $P(x,t)$, the observation density $g$ follows a normal distribution, i.e. $g\sim N(\hat{y_t},\sigma^2)$.

\subsection{Prior injection}

To overcome limits of current music information retrieval systems, a practical engineering solution was to use computational techniques from statistics and digital signal processing allowing the insertion of prior knowledge from different scientific disciplines (e.g. cognitive science, neuroscience, musicology, musical acoustics) \citep{Engelmore1988,Nawab1992,Carver1992,Ellis1996}. Such systems perform a process of reconciliation between the observed acoustic features and the predictions of an internal model of the data-producing entities in the environment. This approach is close to human experience, who perceive the sound in view of different hierarchical levels of prior knowledge, using a collection of global properties, such as musical genre, tempo, and orchestration, as well as more instrument specific properties, such as timbre. It has been widely applied in many music information retrieval tasks, such as genre recognition and automatic music transcription \citep{Ellis1996,Godsmark1999,Bello2000c,Ryynanen2004,Klapuri2004b,Benetos2013c}. 

In mathematical terms, priors are used to sharpen up estimation of model parameters by emphasizing the most likely values in their distributions. This prior integration is performed during state generation, by re-weighting each particle value with a corresponding prior gain. Adding prior knowledge on parameters $p_t$  leads up to sample from the posterior distribution,

\begin{equation}\label{PriorComput}
P(p_t|y_t)\sim P(y_t|p_t)P(p_t)
\end{equation}

where $P(y_t|p_t)$ identifies to the observation density $g$ and $P(p_t)$ the prior knowledge. When the prior and the likelihood are conjugate, sampling from the posterior distribution is rather straightforward. When the posterior does not have a well known form, as it is the case in most real-life applications, computational statistics methods can be introduced to sample from the posterior. The Metropolis-Hasting algorithm \citep{Roberts1997,Newman1999,Robert1999}, based on Monte Carlo methods, brings a powerful framework to tackle that issue. This algorithm is a random walk that uses an acceptance/rejection rule to converge to the specified target distribution, and proceeds as described in the algorithm \ref{Algo_MH}.

\begin{center}
\begin{algorithm}
\centering 

\caption{Metropolis-Hasting algorithm for prior integration.}\label{Algo_MH}
\begin{algorithmic}[1]

\State Draw a starting point $p_t^0$, for which $P(p_t^0|y_t)>0$, from a starting distribution $p_0(p_t)$ ;

\For{$q=1,2,\ldots$}

\State Sample a proposal $p_t^*$ from a jumping distribution at iteration $q$, $J_q(p_t^*|p_t^{q-1})$ ;

\State Calculate the ratio of densities,

		\begin{equation}
			r=\frac{P(p_t^*|y_t)/J_q(p_t^*|p_t^{q-1})}{P(p_t^{q-1}|y_t)/J_q(p_t^{q-1}|p_t^{*})}
		\end{equation}

\State Set

		\begin{equation}
			p_t^q=
			\left \{ 
			\begin{array}{l c r}
				p_t^*& \text{with probability }\min(r,1)\\
				p_t^{q-1}& \text{otherwise}
			\end{array}
			\right .
		\end{equation}
		
\EndFor

\end{algorithmic}
\end{algorithm}
\end{center}

Considering the filtering particle framework defined above, few remarks about the different steps can be highlighted. First, the initial draw is replaced by the PF draw we want to interfere in. Even if the posterior distribution is unknown, the ratio $r$ can be computed as the ratio of the product of the likelihood and the prior since the normalization constant is removed in the ratio. 

\begin{equation}\label{ratio_r}
	r=\frac{g(y_t|p_t^*)p(p_t^*)/J_q(p_t^*|p_t^{q-1})}{g(y_t|p_t^{q-1})p(p_t^{q-1})/J_q(p_t^{q-1}|p_t^{*})}
\end{equation}

The jumping distribution $J_q$ is chosen as a normal distribution to simplify the ratio computation. Indeed, the symmetry property of the normal distribution involves that $J_q$ can be removed in eq. \ref{ratio_r}, to get

\begin{equation}
	r=\frac{g(y_t|p_t^*)p(p_t^*)}{g(y_t|p_t^{q-1})p(p_t^{q-1}))}
\end{equation}

\section{Application to AMT}

We now propose an application of our PLCA-PF framework to the task of Automatic Music Transcription (AMT), and present evaluation results on this task with quantitative comparisons with other state-of-the-art methods.

\subsection{Background on AMT}

Work on AMT dates back more than 30 years, and has known numerous applications in the fields of music information retrieval, interactive computer systems, and automated musicological analysis \citep{Klapuri2004b}. Due to the difficulty in producing all the information required for a complete musical score, AMT is commonly defined as the computer-assisted process of analyzing an acoustic musical signal so as to write down the musical parameters of the sounds that occur in it, which are basically the pitch, onset time, and duration of each sound to be played. This task of ``low-level" transcription, to which we will restrict ourselves in this study, has interested more and more researchers from different fields (e.g. library science, musicology, machine learning, cognition), and been a very competitive task in the Music Information Retrieval community \citep{Mirex2011} since 2000. Despite this large enthusiasm for AMT challenges, and several audio-to-MIDI converters available commercially, perfect polyphonic AMT systems are out of reach of today's technology \citep{Klapuri2004b,Benetos2013c}. To overcome these limitations, a practical engineering solution was to use computational techniques from statistics and digital signal processing allowing the insertion of prior knowledge from cognitive science, musicology and musical acoustics \citep{Engelmore1988,Ellis1996}. This approach is close to human experience, in which the perception of sounds is embedded with prior knowledge, using a collection of global properties such as musical genre, tempo, and orchestration, as well as more specific properties, such as the timbre of a particular instrument.

\subsection{Acoustic modeling}

\subsubsection{PLCA formalization}

In the audio framework, PLCA views the input magnitude spectrogram of a sound source as a histogram of ``sound quanta" across time and frequency, and modeling it as as a linear combination of spectral vectors from a dictionary \citep{Smaragdis2006}. PLCA method is then based on the assumption that a suitably normalized magnitude spectrogram, V, can be modeled as a joint distribution over time and frequency, $P(f,t)$, with f is the log-frequency index and t the time index. This quantity can be factored into a frame probability $P(t)$, which can be computed directly from the observed data (i.e. energy spectrogram), and a conditional distribution over frequency bins $P(f|t)$, as follows

\begin{equation}
P(f,t) =  P(t)P(f|t) 
\end{equation}

Spectrogram frames are then treated as repeated draws from an underlying random process characterized by $P(f|t)$. We can model this distribution with a mixture of latent factors related to polyphonic music transcription of single instruments as follows:

\begin{equation}\label{EqPLCA}
P(f|t) =  \sum_{i,m} P(i|t) P(m|i,t) P(f|i,m) 
\end{equation}

where $P(f|i,m)$ are the spectral templates for pitch $i \in\ \textbf{I}$ (with \textbf{I} the set of pitches, and $N_I$ the number of pitches) and playing mode $m \in\ \textbf{M}$ (with \textbf{M} the set of playing modes, and $N_m$ the number of modes), $P(m|i,t)$ is the playing mode activation, and $P(i|t)$ is the pitch activation (i.e. the transcription). In this paper, the playing mode m will refer to different dynamics of instrument playing (i.e. note loudness). \citep{Smaragdis2008} extended the PLCA model of eq. \ref{EqPLCA} by exploiting the fact that in a CQT, a change of fundamental frequency is reflected by a simple frequency translation of its partials, resulting in a shift invariance over log-frequency. The model proposed, called Shift-Invariant PLCA (SIPLCA), then consists in shifting the templates $P(f|i,m)$ over the log-frequency range of the CQT, thus performing a multi-pitch detection with a frequency resolution higher than MIDI scale. Eq. \ref{EqPLCA} is re-written as follows

\begin{equation}\label{Eq_SIPLCA}
	P(f|t)= \sum\limits_{i,\delta_f,m} P(i|t) P(m|i,t) P(\delta_f | i,t) P(f-\delta_f | i,m)
\end{equation}

where $\delta_f$ is the pitch shifting factor. To constrain $\delta_f$ so that each sound state template is associated with a single pitch, the shifting occurs in a semitone range around the ideal position of each pitch. Thus because we are using in this paper a log-frequency representation with a spectral resolution of 60 bins/octave, i.e. a 20 cent resolution, we have $\delta_f$ $\in$ [-2:2], with $N_{\Delta_f}$ the length of this set of values. 

\begin{equation}\label{Eq_SIPLCA_PFarg}
	P(f,t) = \sum\limits_{i,m,\delta_f}A_t(i)B_t(i,m)C_t(i,m,\delta_f)P(f-\delta_f|i,m)
\end{equation}

In eq. \ref{Eq_SIPLCA_PFarg}, we also identify the different PF arguments, where at a given time $t$, $A_t$ is a vector of length $N_I$ representing the elements of the pitch activity matrix $P(i,t)$ (equal to $P(t)P(i|t)$, through the Baye's rule), $B_t(i,s)$ is the $N_I \times N_m$ matrix whose coefficients are the weights $P(s|i,t)$, and $C_t$ is the $N_I \times N_m \times N_{\Delta_f}$ tensor corresponding to the spectral weights coefficients $P(\delta_f|i,t,m)$. The spectral shifted templates $P(f-\delta_f|i,m)$ are extracted from isolated note samples using a one component PLCA, and are not updated. 

Eventually, as in most spectrogram factorization-based transcription or pitch tracking methods \citep{Grindlay2011,Mysore2009,Dessein2010}, we use a simple threshold-based detection of the note activations from the pitch activity matrix $P(i,t)$, followed by a minimum duration pruning. The threshold for minimum duration for pruning was set to 50 ms. The use of this simple thresholding method should allow one to better highlight the intrinsic differences from the different AMT systems we will compare.

\subsubsection{Particle filter argument}

Since, the spectral shifted templates $P(f-\delta_f|i,m)$ are learned, the set of unknown parameters is $\{A_t,B_t,C_t\}$, and $y_t=V_{ft}(.,t)\in [0,1]^F$ denotes the observations. As described in \ref{transidensity}, each marginal vector $A_t(s)$, $B_t(s)$ and $C_t(s,z)$ are independently forecasted through a Dirichlet distribution, as follows

\begin{equation}
	A_t \sim Dir(\theta_t^1,\ldots,\theta_t^I)
\end{equation}
\begin{equation}	
	B_t(s)\sim Dir(\delta_1(s)_t^1,\ldots,\delta_1(s)_t^I)
\end{equation}
\begin{equation}		
	C_t(s,\delta_f) \sim Dir(\delta_2(s,\delta_f)_t^1,\ldots,\delta_2(s,\delta_f)_t^I)
\end{equation}

Concerning the $\phi$, $\psi_1$ and $\psi_2$ distributions producing the states $(\theta,\delta_1,\delta_2)$, we opt for a Gamma distribution to obtain a non biased transition and to control the variance of the transition. That choice leads to the following transition rules where $\forall i,s,\delta_f$,

\begin{equation}
	\theta_{t+1}^i =\theta_{t}^i\times \alpha_t^i,\alpha_t^i\sim \Gamma(a^i,b^i) 
\end{equation}
\begin{equation}		
	\delta_1(s)_{t+1}^i =\delta_1(s)_{t}^i\times \gamma_t^i,\gamma_t^i\sim \Gamma(c_{s}^i,d_{s}^i)
\end{equation}
\begin{equation}	
	\delta_2(s,\delta_f)_{t+1}^i =\delta_2(s,\delta_f)_{t}^i\times \lambda_t^i,\lambda_t^i\sim \Gamma(e_{s,\delta_f}^i,f_{s,\delta_f}^i)
\end{equation}

with hyperparameters $a^i$, $b^i$, $c_{s}^i$, $d_{s}^i$, $e_{s,\delta_f}^i$ and $f_{s,\delta_f}^i$. Conditionally to $\theta_t$ and $\beta_t$

\begin{equation}
	\theta_{t+1}^i\sim\Gamma(a_s^i,\frac{b_s^i}{\theta_{t}^i})
\end{equation}
\begin{equation}	
	\delta_1(s)_{t+1}^i\sim\Gamma(c_{s}^i,\frac{d_{s}^i}{\delta_1(s)_{t}^i})
\end{equation}
\begin{equation}	
	\delta_1(s,\delta_f)_{t+1}^i\sim\Gamma(e_{s,\delta_f}^i,\frac{f_{s,\delta_f}^i}{\delta_2(s,\delta_f)_{t}^i})
\end{equation}

\begin{figure}[t]
  \centering
  \centerline{
  \includegraphics[width=0.8\columnwidth]{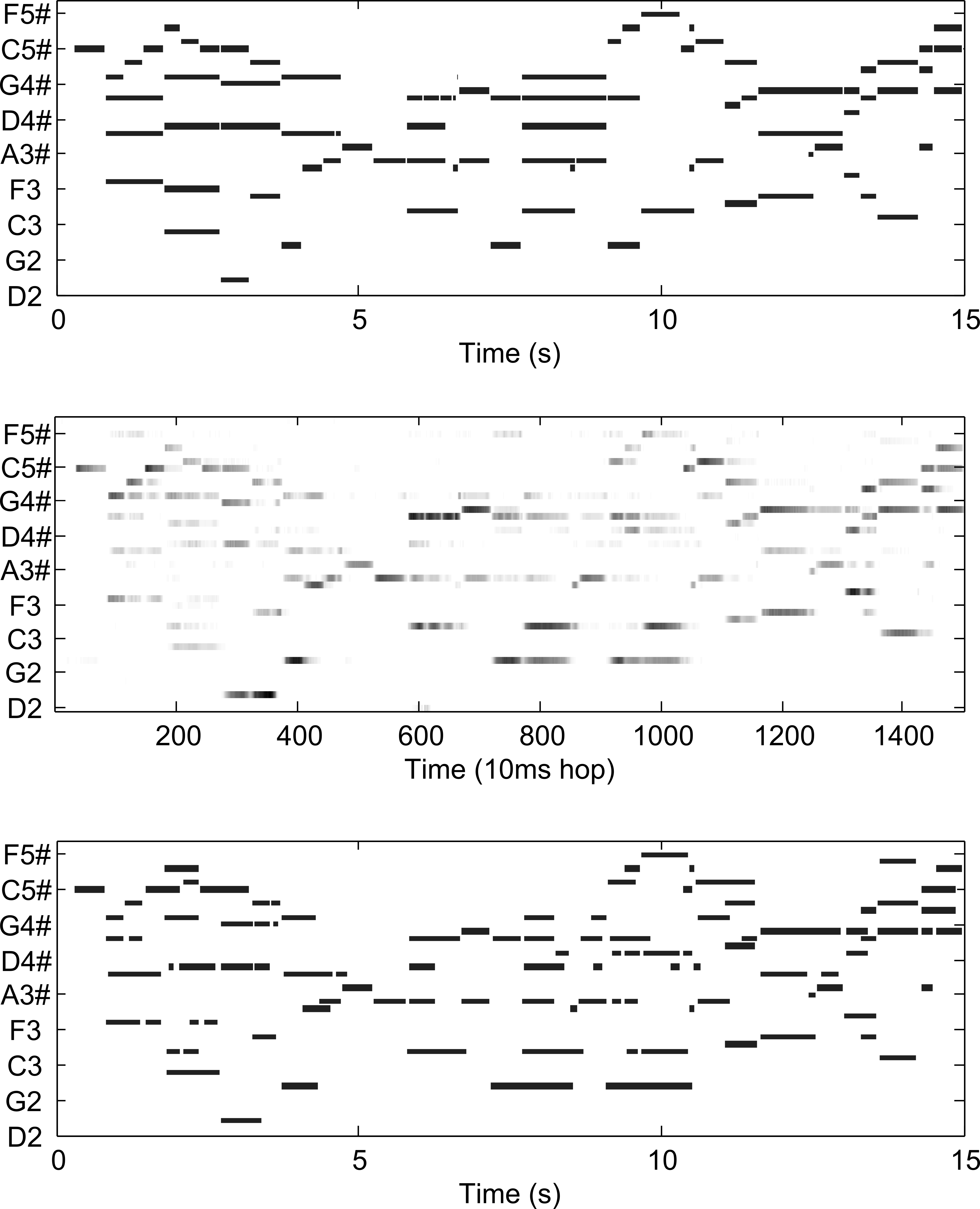}
  }
  \caption{Illustration of different stages of our PF-PLCA system on a test musical sequence, with from top to bottom: ground truth, pitch activity matrix P(i,t) and piano-roll transcription output.}
  \label{MPEAllGraphs}
\end{figure}

Recalling that the mean of $\theta_{t+1}^i$  is $\frac{a_s^i\theta_{t}^i}{b_s^i}$, a non biased transition for $\theta$ involves that $a_s^i={b_s^i}$ since $E(\theta_{t+1}^i)=\theta_{t+1}^i$ is expected. Under the same argument, $c_{s}^i={d_{s}^i}$ and $e_{s,\delta_f}^i=f_{s,\delta_f}^i$. Figure \ref{MPEAllGraphs} provides an example of piano-roll transcription output obtained with our PF-PLCA system.

\subsection{Prior knowledge integration}

A musical signal is highly structured, in both time and frequency domains. In time domain, tempo and beat specify the range of likely note transition times. In the frequency domain, as audio signals are both additive and oscillatory (musical objects in polyphonic music superimpose and not conceal each other), several notes played at the same time form chords, or polyphony\footnote{Here polyphonic music refers to a signal where several sounds occur simultaneously. Whereas in monophonic signals, at most one note is sounding at a time.}, merging their respective spectral structures. When designing priors for an AMT system, one basically aims to help the system figuring out "which notes are present at time t" and "by which ones they will be followed". These two types of information belong respectively to frame-wise spectral priors (e.g. sparseness, spectrum modeling including inharmonicity \citep{Rigaud2013b}) and to frame-to-frame temporal priors (e.g. harmonic content transitions, smoothing of spectrum envelop), and will both be developed into our PLCA-PF framework.

In transcription systems with a general application \citep{Emiya2010,Fuentes2013,Benetos2013}, prior knowledge is generally incorporated with no regard to their musical/physical sense, but with the sole preoccupation of convergence optimization and enhancement of transcription results on a specific musical corpus. As a consequence, priors mostly take the form of a single constant factor, set arbitrarily after simulation experiments. Here, we propose a more complex modeling of musical signal with explicit music-related knowledge. 

To do so, we quantify relations of influence between the different pitches of the instrument pitch range, either within a frame (for spectral priors), or from one frame to the next one (for temporal priors). These priors will then take the form of a matrix S of size $N_I^2$, which quantifies average relations of influence between the $N_I \times N_I$ couples of different pitches, and is defined as follows, $\forall (i,j)\in \{1,\cdots,N_I\}^2$,

\begin{equation}\label{Symp}
 	S=  \begin{pmatrix}
	S(1,1)&\cdots&S(1,N_I) \\
	\vdots& S(i,j) &\vdots\\
	S(N_I,1)&\cdots&S(I,N_I)
	\end{pmatrix}
\end{equation}

The PLCA-PF framework allows a general insertion of this matrix through the term $P(p_t)$ of eq. \ref{PriorComput}, to which we can give the following form

\begin{equation}\label{symprior}
	P( p_t )\propto \exp(-p_t' S K_{p_t})
\end{equation}

where $p_t\in\{A_t,B_t(s),C_t(s,\delta_f)\}$, $\forall (s,\delta_f)\in\{1,\ldots,S\}\times\{1,\ldots,\Delta_f\}$, and $K_{p_t}$ is a vector of length $N_I$ associated to $p_t$. 

It is noteworthy that simpler modeling of prior knowledge, such as a simple pitch-dependent vector, can also take the form of a diagonal matrix S of size $N_I^2$, with the vector values put into this diagonal (the zero-coefficients of S provide an unitary prior value which does not affect particle weights).

\subsubsection{Sparse priors}

During the multi-pitch estimation step of an AMT process, a too much large number of non-zero activation scores is often observed, making the operation of ``finding the right notes" more difficult. In order to overcome this flaw, a sparseness prior can reduce the number of active notes per frame in selecting the most salient ones. Previous works mostly use pitch-independent sparse prior in PLCA-EM algorithms. \citep{Fuentes2013} compute a sparseness prior $P(A_t)$ to constrain the impulse distribution $A_t$, as follows

\begin{equation}
	P(A_t) \propto \exp\left({-2\beta\sqrt{J}||A_t)||_{1/2}}\right)
\end{equation}

with $||A_t||_{1/2}=\sum\limits_i\sqrt{A_t(i)}$ and $\beta$ a positive hyperparameter indicating the strength of the prior. With this prior, a numerical fixed point algorithm is required to obtain a solution with the EM  algorithm. Other works \citep{Grindlay2011,Benetos2011,Benetos2013} impose sparsity on the pitch activity matrix and the pitch-wise source contribution matrix by modifying EM equations. 

The common point to all these EM-based sparse priors is that they are pitch-independent, and rely on hyper-parameters, which are either arbitrary set and/or optimized on a given sound dataset. In this paper, we define sparse priors informed by explicit musical acoustics related knowledge. Musically, the occurrence of simultaneous notes can result either from ``acoustic polyphony", or from ``musical polyphony". "Acoustics polyphony" is strongly related to the timbre of the instrument, and more precisely to the physical phenomena of mutual resonances and note persistence. Although this type of polyphony is an integral part of instrument timbre, it represents a noise signal added to the actual played note from the point of view of music transcription. For what concerns ``musical polyphony", it corresponds to the note combinations played by the musician and intended by a composer with a proper polyphonic writing. It directly provides useful information about which notes are commonly played simultaneously in a musical piece. Prior knowledge can be learned from both of these polyphonic origins for an instrument repertoire, studying respectively the timbre of the instrument or the frame-wise musical characteristics of the repertoire.

A first sparse prior $P_{spa1}$ on note mixture likelihood has then been defined. Following the form of matrix S (eq. \ref{Symp}), each coefficient is computed by a frame-wise counting of the pitches j played simultaneously to pitch i, from our training MIDI transcripts (see Sec. \ref{SoundData} for details on the sound database). We propose a second sparse prior $P_{spa2}$ on mutual resonances. For strings on a bowed, plucked, or hammered instruments, mutual resonances result from sympathetic strings, which vibrate (and thereby sound a note) in sympathetic resonance with the note sounded near them by some other agent. Here, to compute each coefficient $S(i,j)$, we used two datasets of isolated notes, a first one composed of free-resonating notes, and a second one in which all strings were muted excepting the played one. For each note sample of these two datasets, the spectrum was computed with a FFT using a 4096-sample Hamming window after the onset, unitary normalized and labelled $X_{(d,i)}$ for pitch i and dataset d (equal to 1 or 2). We then used the algorithm \ref{AlgoSij} to get the scores S(i,j).

\begin{algorithm}
\centering 

\caption{Computation of coefficients S(i,j).}\label{AlgoSij}
\begin{algorithmic}[1]

\For{For each pitch i $\in$ I}

\State $\tilde{X}_i$ = $|| X_{(1,i)} - X_{(2,i)} ||$

\State Binary thresholding of $\tilde{X}_i$, i.e.

\vspace{-0.5cm}

\[ \tilde{X}_{i}(f) = \left\{ 
  \begin{array}{l l}
    1 & \quad \text{for $f =arg(\tilde{X}_i \ge 0.5)$}\\
    0 & \quad \text{otherwise}
  \end{array} \right.\]

\vspace{-0.2cm}

\For{Each pitch j $\in I , j \neq i$}

\State S(i,j) = $X_{(2,j)} \cdot \tilde{X}_{i}$, with [$\cdot$] the element-wise product

\EndFor

\EndFor

\end{algorithmic}
\end{algorithm}

\begin{figure}[htbp]
\begin{center}
\includegraphics[width=0.45\columnwidth]{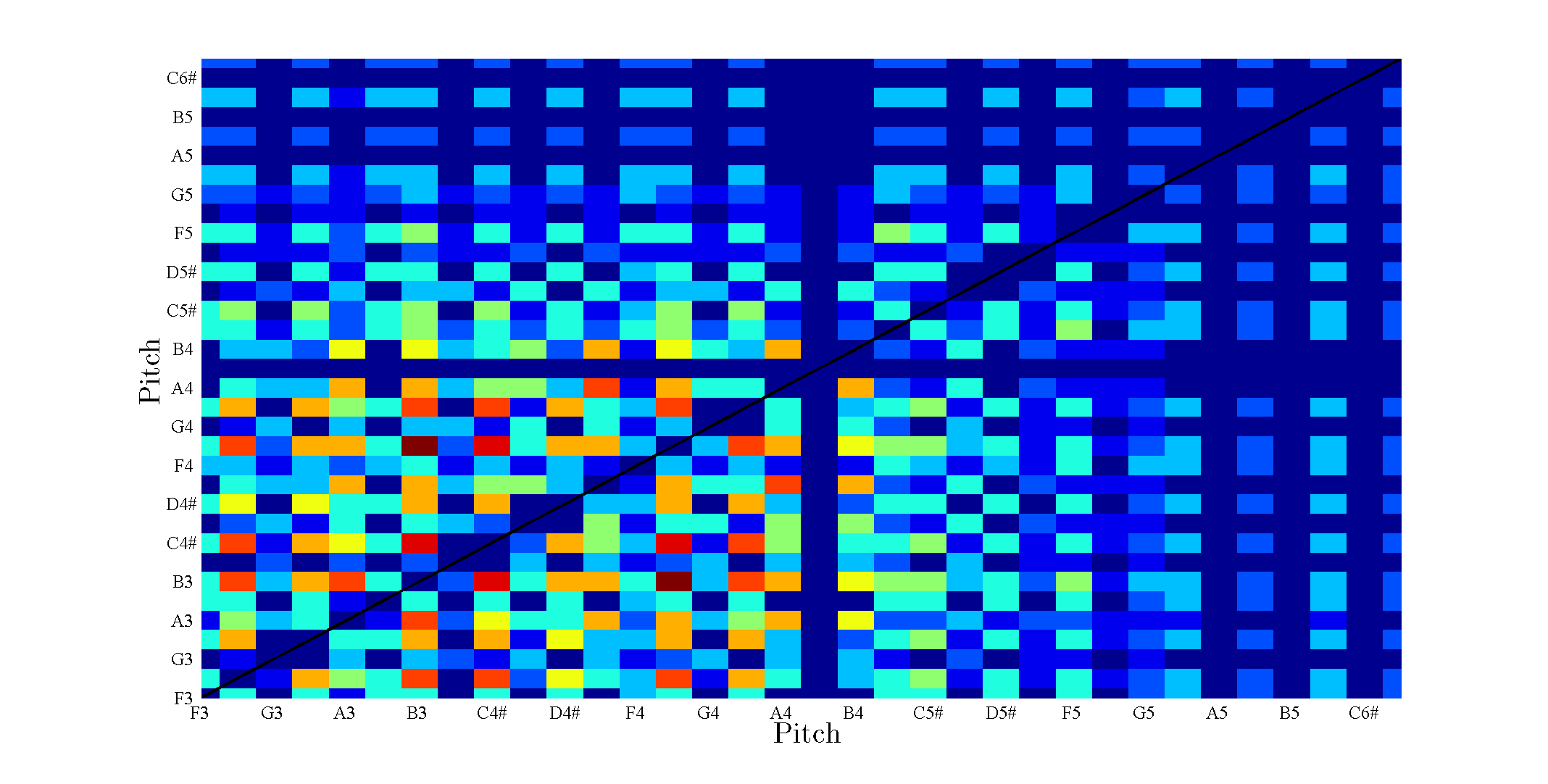}
\includegraphics[width=0.45\columnwidth]{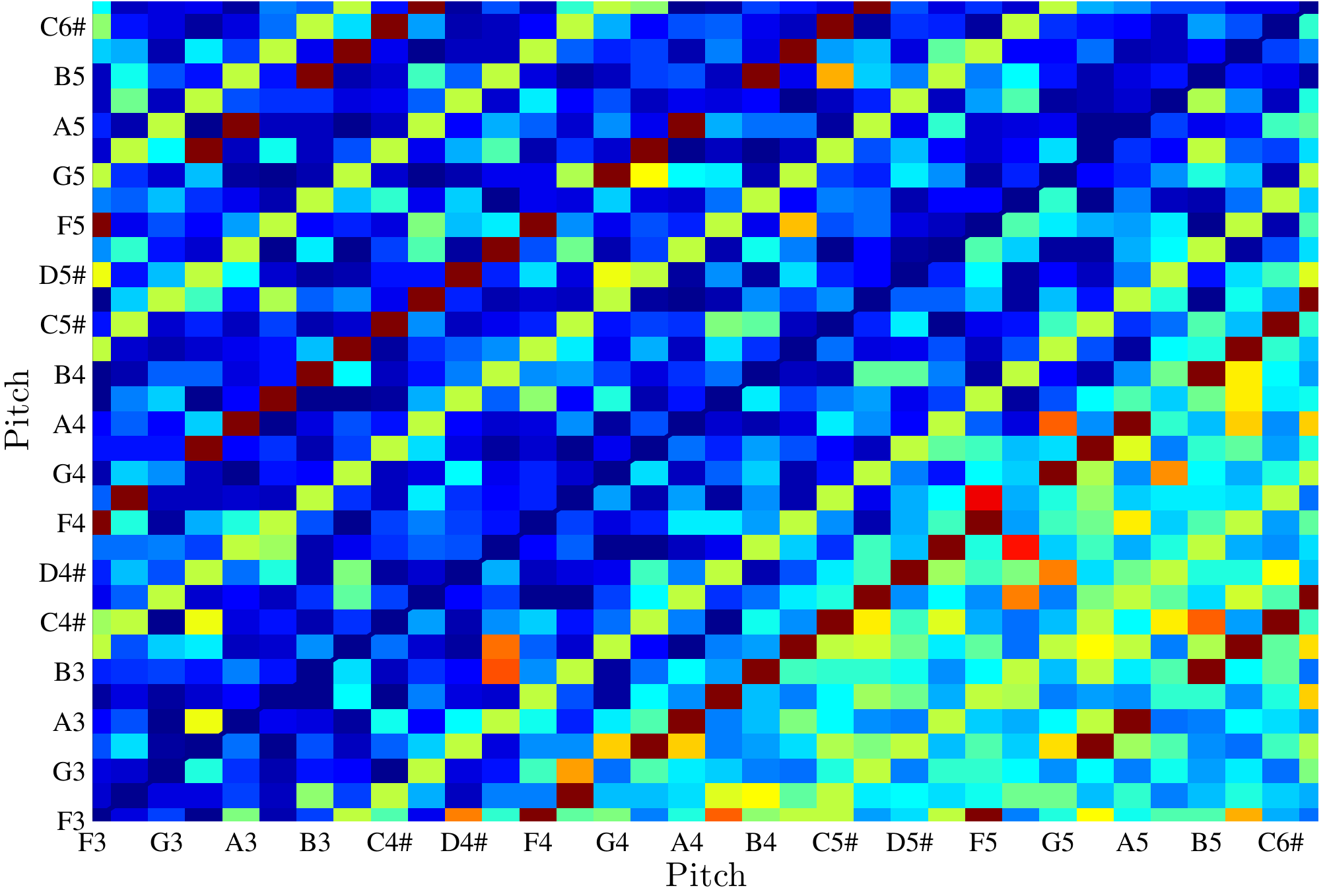}
\caption{Illustration of the inter-pitch influence matrix S for chord content (on the left) and sympathetic resonances (on the right).}
\label{AllPriorMatrices}
\end{center}
\end{figure}

These two priors $P_{spa_1}$ and $P_{spa_2}$ are represented in figure \ref{AllPriorMatrices} through their respective matrices $S_{spa_1}$ and $S_{spa_2}$. Eventually, for this sparse type prior, the set of prior parameters $p_t$ in eq. \ref{symprior} is equal to $A_t$ and $K_{p_t}$ is set to $A_t$ , which becomes
 
\begin{equation}\label{Pspa}
	P_{spa}(A_t)\propto \exp(-A_t' S_{spa} A_t)
\end{equation}

For prior $P_{spa1}$, before injecting it in eq. \ref{Pspa}, we normalized its matrix $S_{spa1}$ with the operator $\Pi$ (eq. \ref{EqPi}) defined as

\begin{equation}\label{EqPi}
\Pi(x) = 1 - \frac{x}{max(x)}
\end{equation}

as we need knowledge rejecting the hypothesis of certain pitch combinations. 

%It is noteworthy that the authors \citep{Cazau2015c} have already developed similar pitch-dependent sparse priors through the algorithm EM using a fixed point algorithm, as there is no closed-form solution for each $A_t(j)$. A solution can then be found by increasing the posterior probability at each iteration of the EM algorithm, with the following EM update rule, where $\lambda$ is the Lagrangian multiplier coefficient, $\forall j\in \{1,\cdots,I\}$,
%
%\begin{equation}\label{sympestim}
%\hat{A_t}(j)=\frac{w_j}{\sum\limits_{i=1}^I S_{sr}(ij) \hat{ A_t}(i)+\lambda}
%\end{equation}

%, by simply counting the occurrences of each chord and then normalizing over each pitch to provide probabilities, i.e. $\sum_{i=I} S_{i,j} = 1, \forall j \in I$. 

\subsubsection{Sequential priors on harmonic transitions}

Many previous works \citep{Poliner2007,Grindlay2011,Benetos2013} on AMT have used sequential priors to model each pitch activity/inactivity phases, which is done using two-state on/off HMMs for each of them during a post-processing stage. This operation performs a time filtering of note detection decision, which mainly avoids a lot of single miss errors and smooths note boundaries. But musically, the information is very restricted, as it consists only in knowing how long a given pitch note remains active, which can result from both playing techniques of the musician and vibratory properties of the instrument. 

The sequential prior we present in this paper is defined as the probability to switch between two successive mixtures of notes in a musical piece. These transition probabilities are determined by sampling the training MIDI transcripts at the precise times corresponding to the analysis frames of the activation matrix, and just checking for the presence of a note in each frame. These probabilities give us a global view of the usual and unusual harmonic transitions of an instrument repertoire. Original mixtures, i.e. those not encountered during the training phase, get a likelihood weighted accordingly to the Witten-Bell discounting algorithm \citep{Witten1991}. This prior $P_{tra}$ is represented in figure \ref{AllPriorMatrices} through their matrix $S_{tra}$. Eventually, for this sequential type prior, $p_t$ in eq. \ref{symprior} is equal to $A_t$ and $K_{p_t}$ is set to $A_{t-1}$, which leads to

\begin{equation}
P_{tra}(A_t) \propto \exp(-A_t' S_{tra} A_{t-1})
\end{equation}

\subsubsection{Prior combination}

The PLCA-PF framework offers an easy-to-implement unifying way of integrating priors from both time and frequency domains. In this framework, priors are injected during the filtering process through eq. \ref{PriorComput}, and modify the parameters without disturbing their generation. In the set of parameters $p_t$, the independence between each parameter $p_t^n$ leads to 

\begin{equation}\label{intra}
P(p_t) \propto \prod\limits_{n} P(p_t^n)
\end{equation}

Within a defined parameter $p_t^n$, the general prior $P(p_t^n)$ can be seen as the product of the different priors $P_{prior}^n$ associated to $p_t^n$ 
%\begin{equation}
%p(p_t) \propto \prod\limits_{n}\prod\limits_{prior}p_{prior}(p_t^n))
%\end{equation}

\begin{equation}\label{inter}
P(p_t^n) \propto \prod\limits_{prior} P_{prior}^n(p_t^n)
\end{equation}

Using equations \ref{symprior}, \ref{intra} and \ref{inter}, we combine the different priors, characterized by their respective matrices $S_{prior}$, as follows

\begin{equation}
P(p_t) \propto \exp(- \sum\limits_n \sum\limits_{prior} {p_t^n}' S_{prior}^n K_{prior}^n)
\end{equation}

%\begin{equation}
%p(p_t) \propto \prod\limits_{prior}\exp(-p_t' S_{prior} K_{prior})
%\end{equation}

%\begin{equation}
%p(p_t) \propto \exp(-p_t '\sum\limits_{prior}S_{prior} K_{prior})
%\end{equation}

%\begin{align}
%p(A,B,C)& \propto p(A)p(B)p(C)\\
%& \propto \prod_i p_i(A)\prod_j p_j(B) \prod_kp_k(C)\\
%\end{align}
%1ere ligne : independence intra (entre les différents paramètres)\\
%2eme ligne : indépendance inter (au sein d'un même à priori\\

%\textcolor{red}{Contrary to EM estimation, an infinite number of generic priors can be added this way since no log posterior distribution needs to be maximized.}

\subsection{Practical implementation}

As a time-frequency representation, all input signals sampled undergo a Q-constant with 60 bins/octave, with window size of 23 ms (1024 coefficients at 44.1-kHz sampling rate) and a 50 \% hop, which is adequate for the tonal part of the signal. We now present the different algorithms implemented in our PF framework.

\subsubsection{Filtering algorithm}

Such as in \citep{Fong2002}, we develop a generic PF algorithm assuming that the proposal distribution $q(x_t|x_{1:t-1}^{(i)},y_{1:t})=f(x_t|x_{t-1}^{(i)})$, as detailed in the pseudo-algorithm \ref{GenPF}. The resample step is processed as detailed in the pseudo-algorithm \ref{ResampAlgo}.

\begin{center}
\begin{algorithm}
\centering 

\caption{Generic particle filtering with multinomial resampling}\label{GenPF}
\begin{algorithmic}[1]

\State Let $f(x_1|x_0)=f(x_1)$ be state prior distribution. Then for $t=1$ to $T$ :

\State $\forall i\in \{1,\cdots,N\}$, generate $N$ samples from the proposal $q(x_t|x_{1:t-1}^{(i)},y_{1:t})=f(x_t|x_{t-1}^{(i)})$,
\begin{center}
	$x_t^{(i)}\sim f(x_t|x_{t-1}^{(i)})$
\end{center}

\State $\forall i\in \{1,\cdots,N\}$, evaluate the importance weights and normalise
\begin{center}
	$w_t^{(i)}\propto g(y_t|x_t^{(i)}),\sum\limits_{i=1}^Nw_t^{(i)}=1$
\end{center}

\State Resample $\{x_t^{(i)};i=1,\cdots,N\}$ $N$ times with replacement.

\end{algorithmic}
\end{algorithm}
\end{center}

\begin{center}
\begin{algorithm}
\centering 

\caption{Multinomial resampling}\label{ResampAlgo}
\begin{algorithmic}[1]

\State Initialize the CDF : $c_1=0$
\State $\forall i\in \{2,\cdots,N\}$
	\begin{itemize}
		\item Construct CDF : $c_i=c_{i-1}+w_t^{(i)}$
	\end{itemize}
\State Start at the bottom of the CDF : $i=1$
\State Draw a starting point : $u_1\sim U[0,\frac{1}{N}]$
\State $\forall j\in \{1,\cdots,N\}$
\begin{itemize}
	\item Move along the CDF : $u_j=u_1+\frac{j-1}{N}$
	\item While $u_j>c_i$, $i=i+1$
	\item Assign sample : $x_t^{(j)*}=x_t^{(i)}$
	\item Assign weight : $w_t^{(j)}=1/N$
\end{itemize}
\State Return $\{x_t^{(k)*},w_t^{k} \}_{k=1}^{N}$.

\end{algorithmic}
\end{algorithm}
\end{center}

\subsubsection{Smoothing algorithm}

After having generated weighted particles $\{x_t^{(i)},w_t^{(i)};i=1,\cdots,N,t=1,\cdots,T\}$ from the PF, the smoothing algorithm proceeds as detailed in the pseudo-algorithm \ref{SmoothAlgo}.

\begin{center}
\begin{algorithm}
\centering 

\caption{Generic particle smoother}\label{SmoothAlgo}
\begin{algorithmic}[1]

\State Choose $\widetilde{x}_T=x_T^{(i)}$ with probability $w_T^{(i)}$.

\State For $t=T-1$ to 1 :
\begin{itemize}
	\item Calculate $w_{t|t+1}^{(i)}\propto w_t^{(i)}f(\widetilde{x}_{t+1}|x_t^{(i)})$ for each $i=1,\cdots,N$;
	\item Choose $\widetilde{x}_t=x_t^{(i)}$ with probability $w_{t|t+1}^{(i)}$.
\end{itemize}
\State $\widetilde{x}_{1:T}=(\widetilde{x}_1,\widetilde{x}_2,\cdots,\widetilde{x}_T)$ is an approximate realisation from $p(x_{1:T}|y_{1:T})$.

\end{algorithmic}
\end{algorithm}
\end{center}

\subsection{Evaluation procedure}

\subsubsection{Evaluation AMT systems}

To evaluate comparatively transcription performance of our PLCA-PF algorithm, we tested different algorithms on the same test datasound. Table \ref{RecMPEalgos} provides an overview of these algorithms. HALCA is short for the Harmonic Adaptive Latent Component Analysis algorithm\footnote{Codes are available at \url{http://www.benoit-fuentes.fr/}.} \citep{Fuentes2013}. Here, each note in a constant-Q transform is locally modeled as a weighted sum of fixed narrowband harmonic spectra, spectrally convolved with some impulse that defines the pitch. All parameters are estimated by means of the EM algorithm, in the PLCA framework. This algorithm was recently evaluated by MIREX and obtained the $2^{nd}$ best score \citep[$2^{nd}$ best scores in the Multi-Pitch Estimation task, 2009-2012]{Mirex2011}. The algorithm PLCA-EM is the algorithm proposed by \citep{Benetos2013d}, whose main characteristics is its use of pre-defined templates, allowing them to avoid updating the templates in the maximization step of the EM algorithm. This algorithm is also state-of-the-art \citep[$1^{st}$ best scores in the Multi-Pitch Estimation task, 2009-2012]{Mirex2011}. PLCA-DAEM is the same as PLCA-EM, only replacing the EM algorithm by a DAEM algorithm. \citep{Cheng2013} observed significant improvements on their transcriptions through this modification.

\begin{table}[h]
\centering{
\resizebox{8.5cm}{!}{
\begin{tabular}{|c|c|c|c|}
\hline
Method name      & References            & \begin{tabular}[c]{@{}c@{}}Parameter estimation\\ algorithm\end{tabular} & Priors                            \\ \hline
HALCA            & \citep{Fuentes2013}  & EM                                                                       & Sparsity + Continuity + Unimodal  \\ \hline
PLCA-EM          & \citep{Benetos2013d} & EM                                                                       & Sparsity                          \\ \hline
PLCA-DAEM        & \citep{Cheng2013}    & DAEM                                                                     & Sparsity                          \\ \hline
PLCA-PF          & Proposed              & PF                                                                       & --                                \\ \hline
PLCA-PF + priors & Proposed              & PF                                                                       & $P_{spa1}$, $P_{spa2}$, $P_{tra}$ \\ \hline
\end{tabular}
}
}
\caption{Recapitulative table of the AMT systems tested in our simulation experiments.}\label{RecMPEalgos}
\end{table}

\subsubsection{Musical corpus}\label{SoundData}

To test our AMT system and train the sparse priors proposed, we need three different sound corpus: audio musical pieces of an instrument repertoire, the corresponding scores in the form of MIDI files, and a complete dataset of isolated notes for this instrument. We will use two different decay instruments for evaluation, namely the classical piano and the \textit{marovany} zither from Madagascar. For piano, audio data was extracted from the MAPS database \citep{Emiya2010}, which is composed of high-quality note samples and recordings from a real upright piano, whose MIDI scores have been automatically compiled using the Disklavier technology. For the marovany instrument, sound templates and musical pieces were extracted from personal recordings made in our laboratory. Pieces were transcribed with an original multi-sensor retrieval system \citep{Cazau2013c}. 

From these sound databases, we extracted different sets of training and test data, as our prior must be trained using automatically generated knowledge from MIDI files and template datasets. To do so, we first divided each musical pieces into 15-second sequences, which provided us with a total of 1.2 and 0.83 hours of audio, respectively for the piano and marovany datasets. Within each dataset, the musical sequences were randomly split into training and testing sequences, using by default 30 \% of sequences for testing, and the 70\% remaining ones for training. In our simulation experiments, this procedure is repeated five times, and an average is computed on the resulting scores. To prevent any overfitting of our data, we carefully distinguished between training and test data. Especially, sound templates used in the PLCA model were extracted from an instrument model different from the one used in recordings. Also, sequences used to train the priors were not used for evaluation. 

%Also, a musical complexity degree $Q_{mus}$, scaled from 1 to 10, is assigned to each sequence test, based on its polyphonic degree (i.e. the number of simultaneous notes in a time window) and the note debit (i.e. the average number of notes per second).

\subsubsection{Error metrics}

For assessing the performance of our proposed transcription system, we adopt a note-oriented approach, according to which a note event is assumed to be correct if it fills the condition that its onset is within a 50 ms range from a ground-truth onset (i.e. the standard tolerance commonly used \citep{Bello2005,Mirex2011}). Such a tolerance level is considered to be a fair margin for an accurate transcription, although it is far more tolerant than human ears would, as we remind that those are able to distinguish between two onsets as close as 10 ms apart \citep{Moore1997}. Evaluation metrics are defined by equations \ref{Metr1}-\ref{Metr4} \citep{Mirex2011}, resulting in the note-based recall (TPR), precision (PPV) and F-measure (the harmonic mean of precision and recall) :

\begin{equation}\label{Metr1}
\text{TPR} = \frac{\sum_{n=1}^N {\text{TP}[n]}}{\sum_{n=1}^N {\text{TP}[n] + \text{FN}[n]}} \end{equation}
\begin{equation}\label{Metr2}
\text{PPV} = \frac{\sum_{n=1}^N {\text{TP}[n]}}{\sum_{n=1}^N {\text{TP}[n] + \text{FP}[n]}} 
\end{equation}
\begin{equation}\label{Metr4}
\text{F-measure} = \frac{2 . \text{PPV} . \text{TPR}}{\text{PPV}+\text{TPR}} 
\end{equation}

where N is the total number of notes, and TP, FP and FN scores stand for the well-known True Positive, False Positive and False Negative detections. The recall is the ratio between the number of relevant and original items; the precision is the ratio between the number of relevant and detected items; and the F-measure is the harmonic mean between precision and recall. For all these evaluation metrics, a value of 1 represents a perfect match between the estimated transcription and the reference one.

\subsection{Results and discussion}

We present in the following simulation experiments on parameter initialization dependency and transcription performance, comparing our proposed PF-based algorithm with three other state-of-the-art algorithms (see table \ref{RecMPEalgos}).

\subsubsection{Computational time}

\begin{figure}[htbp]
\begin{center}
\includegraphics[width=0.6\columnwidth]{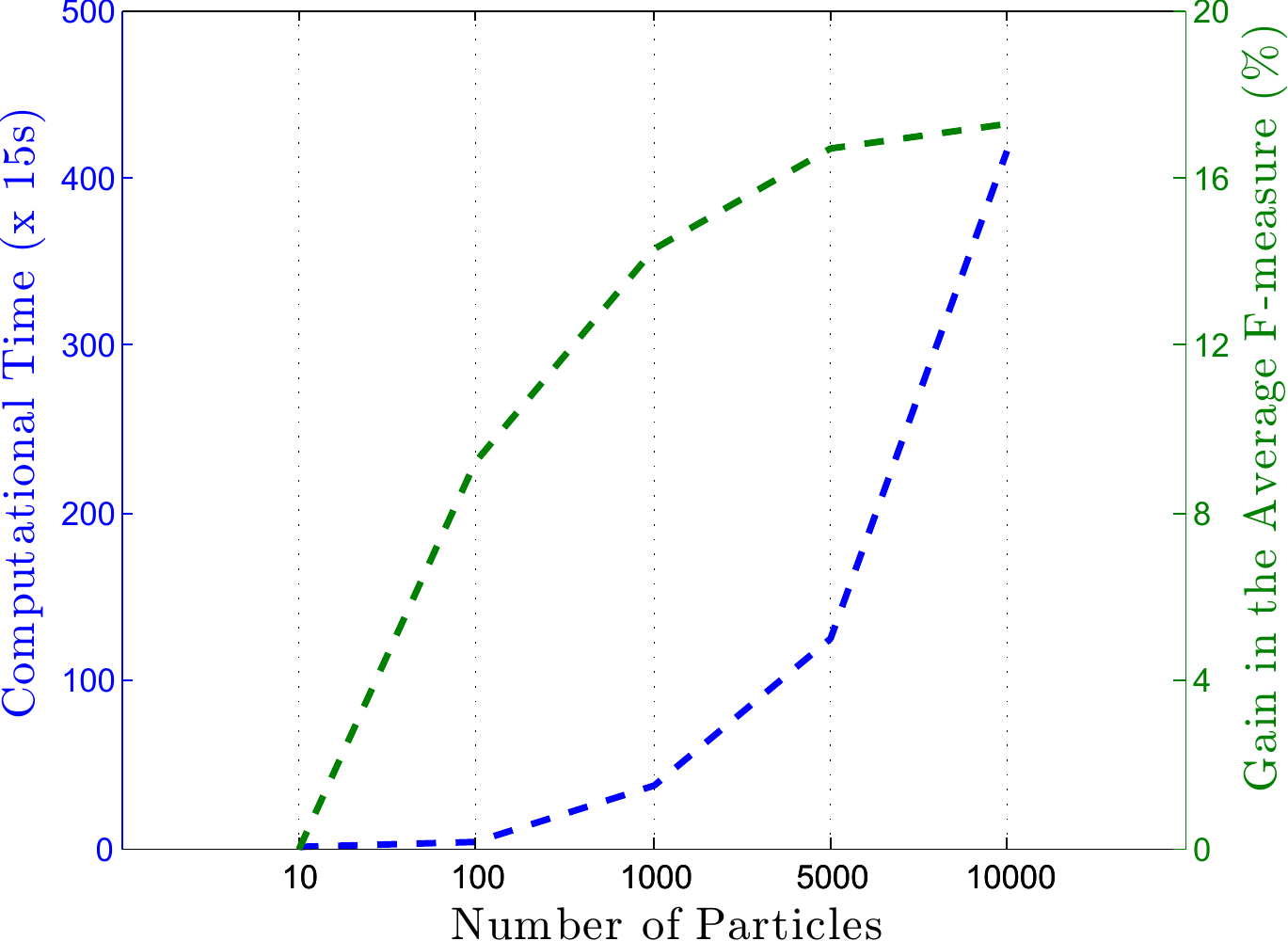}
\caption{Computational time of our PLCA-PF system on a 15-s test musical sequence against the number of particles.}
\label{ComputationalTimeVSNberParticles}
\end{center}
\end{figure}

Figure \ref{ComputationalTimeVSNberParticles} shows the computational time of our PLCA-PF system on a 15-s test musical sequence against the number of particles. Both the computational time and transcription performance increase with the number of particles used. It is well-known that filtering particle is very demanding in computation time, which increases exponentially with the number of particles. 
In its current form, our system is not very efficient computationally, as it produces a transcription in about 50 at real time on a PC computer (e.g., for a 15 sec recording it requires 12.5 mins) with 2000 particles. Increasing the number of particles also increases transcription accuracy, rising the average F-measure with gains as high as 14 $\%$ between 10 and 1000 particles, which begins to stagnate after  5000 particles. This tendency remains observed regardless the instrument repertoire. A trade-off between computation time and transcription precision must then be considered. 

\subsubsection{Parameter initialization dependency}

\begin{figure}[htbp]
\begin{center}
\includegraphics[width=0.8\columnwidth]{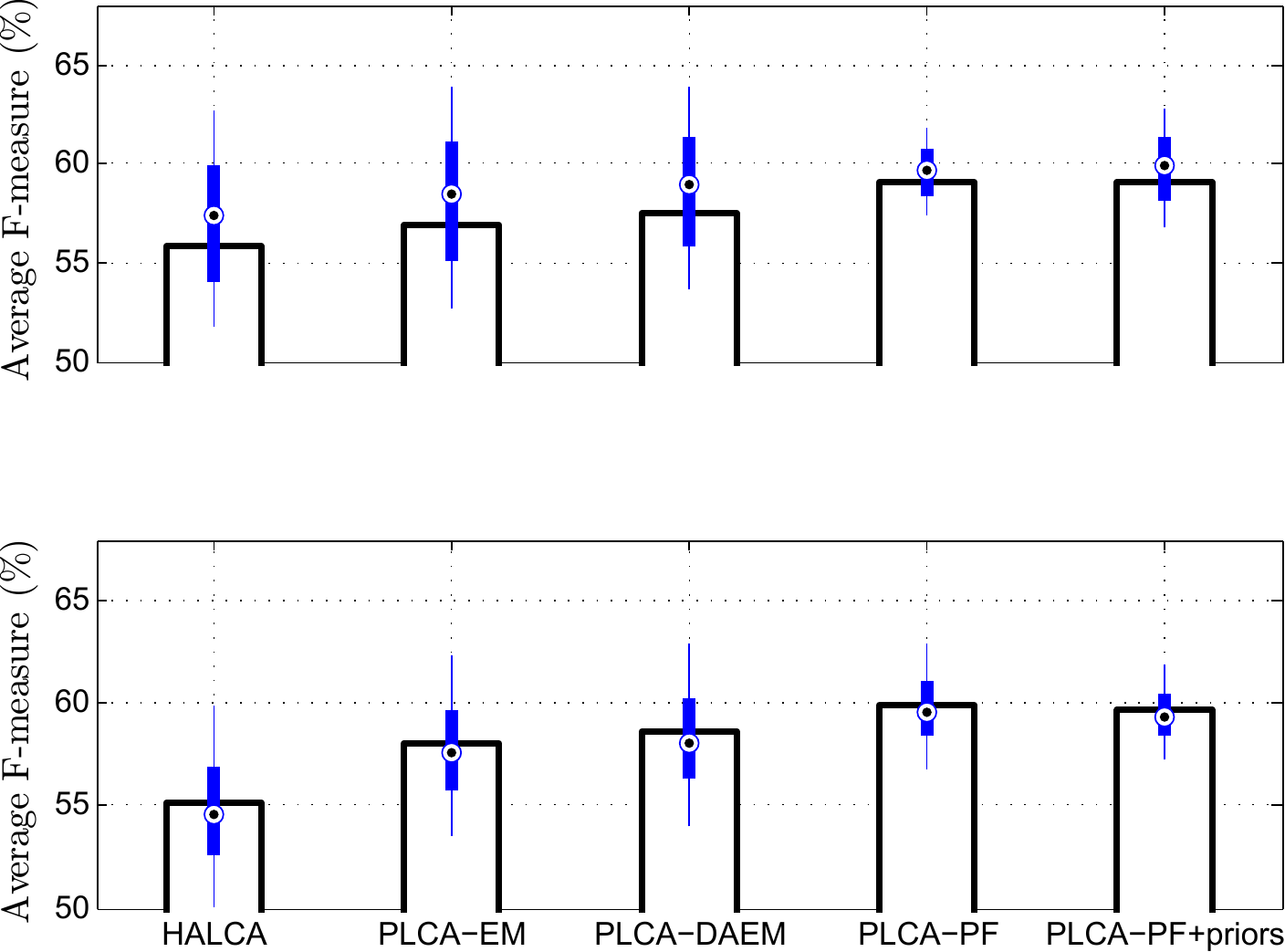}
\caption{Variances in transcription performance using random initialization of system parameters, for the piano (top graph) and the marovany (bottom graph) repertoires. Black bars indicate the reference scores of each system, obtained with an uniform initialization of parameters, as it is commonly done by default (e.g. \citep[Paragraph V.A.1]{Fuentes2013}).}
\label{VariancesTranscriPerfo}
\end{center}
\end{figure}

Figure \ref{VariancesTranscriPerfo} compares dependency of transcription outputs on parameter initialization. These results are computed from 40 simulation trials on one test sequence with a randomized parameter initialization. Black bars indicate the reference scores of each system, obtained with an uniform initialization of parameters, as it is commonly done by default (e.g. \citep[Paragraph V.A.1]{Fuentes2013}). We observe that the proposed systems including filtering particle are globally more robust to parameter initialization, in comparison to other EM or DAEM -based systems, which present an important variability in the average F-measure (e.g., $\pm 2.7 \%$ the PLCA-EM model for the \textit{marovany} repertoire. Then, depending on the sound dataset under evaluation, the set of initialization parameters may be sub-optimal, with performance losses rising as high as 5 $\%$ in the average F-measure. In definitive, making AMT systems less dependent on data should favour their generalization to the diversity of music.

% A current evaluation method commonly used in AMT studies is to keep only the best transcription score over all the test sequences, which contributes to mask the variance of results, in particular the one due to parameter initialization. 

\subsubsection{Transcription performance}

Table \ref{ResultTranscriPerfos} compares transcription performance of our different AMT systems through the different error metrics, for the piano and the \textit{marovany} repertoires, respectively. Both of them present a complex polyphony structure. The \textit{marovany} repertoire is characterized by fast arpeggios, with an ample halo-like sound with rich overtones due to the complex resonating behavior of the instrument. The classical piano repertoire presents more complex and richer chord transitions, with different playing techniques and dynamics which interfere continuously on the timbre of the instrument. For both of these repertoires, the improvements brought by PF-based systems in transcription performance are likely related to the EM limitations evoked in our Introduction, which make EM-based algorithms less efficient in finding active notes in complex polyphonic signals due to problems of local maxima convergences. Also, information from standard deviations shows that our proposed algorithm presents the minimum value for the standard deviation (2.3 $\%$), which implies a higher robustness to transcribe polyphonic signals with different musical features.

\begin{table}[h]
\centering
\resizebox{!}{1.1cm}{
\begin{tabular}{|c|c|c|c|c|c|c|}
\hline
                 & \multicolumn{3}{c|}{Piano} & \multicolumn{3}{c|}{Marovany} \\ \hline
Methods          & TPR   & PPV   & F-measure  & TPR    & PPV    & F-measure   \\ \hline

HALCA            &  55.2  & 59.7  & 57.3      & 55.1 & 57.6 & 56.3 \\ \hline
     
PLCA-EM          & 57.2  & 62.8 &  59.8   & 55.6  & 59.7  & 57.6 \\ \hline

PLCA-DAEM        & 59.1   & 62.9 & 60.9  & 56.1 & 58.3 & 57.2 \\ \hline

PLCA-PF          & 57.9  & 62.2 & 59.9  & 57.7 & 60.1 & 58.8 \\ \hline

PLCA-PF + priors & 61.2 & 62.5 & 61.8    &  58.2 & 60.9 & 59.5   \\ \hline
\end{tabular}
}
\caption{Mean transcription error metrics (in $\%$) for the piano recordings with our different AMT systems.}\label{ResultTranscriPerfos}
\end{table}

% In this section, we test our MPE algorithms on specific difficulties of AMT, commonly evoked in specialized literature, namely octave-related (i.e. simple harmonic relations between simultaneous notes) and sparsity-related (i.e. number of simultaneous notes) problems.

\section{Conclusion}

Current PLCA-based systems for AMT use the well-known EM algorithm to estimate the model parameters. This algorithm presents well-known inherent defaults (local convergence, initialization dependency), making EM-based systems limited in their applications to AMT, particularly in regards to the mathematical form and number of priors. To overcome such limits, we have developed in this paper a different estimation framework based on Particle Filtering, which consists in sampling the posterior distribution over larger parameter ranges. This framework proves to be more robust in parameter estimation, more flexible and unifying in the integration of prior knowledge in the system. It provides the abilities of injecting more complex musicological knowledge, as well as combining simultaneously a theoretically infinite number of priors. Our proposed Particle-Filtering systems achieve promising rankings in terms of accuracy rate, and further experimentations will be necessary to confirm these preliminary results.

\medskip

 \noindent \textbf{Acknowledgements}
 
 \setlength{\parindent}{0.7cm} 
 
At the risk of omitting some relevant names, the authors would like to especially thank March Chemillier (CAMS-EHESS) for his help in recording the \textit{marovany}, and Laurent Quartier (LAM-UPMC) for technical supports.

%
%\bibliography{References_Biblio}
%\bibliographystyle{Cazau}

\end{document}